\documentclass[11pt]{article}

\usepackage[preprint]{latex/acl}

\usepackage{times}
\usepackage{latexsym}

\usepackage[T1]{fontenc}
\usepackage[most]{tcolorbox}
\tcbuselibrary{listings,breakable}
\usepackage{dblfloatfix}

\usepackage[utf8]{inputenc}
\usepackage{booktabs}
\usepackage{float}
\usepackage{multirow}
\usepackage{graphicx}
\usepackage{microtype}
\usepackage{colortbl}
\definecolor{lightblue}{RGB}{220,235,250} % adjust if you want lighter/darker

\usepackage[most]{tcolorbox}
\usepackage{enumitem} % better control of numbering
\usepackage{inconsolata}
\usepackage{amsmath}
\usepackage{subcaption}
\usepackage{amsfonts}

\usepackage{graphicx}

\title{Cross-Lingual LLM-Judge Transfer via Evaluation Decomposition}

\author{%
  \textbf{Ivaxi Sheth$^{1}$\thanks{Work done as an intern at Amazon.} \quad Zeno Jonke$^{2}$ \quad Amin Mantrach$^{2}$ \quad Saab Mansour$^{2}$ }
  \\ \\
 $^{1}$CISPA Helmholtz Center for Information Security \\
 $^{2}$Amazon \\
}

\begin{document}
\maketitle
\begin{abstract}

As large language models are increasingly deployed across diverse real-world applications, extending automated evaluation beyond English has become a critical challenge. Existing evaluation approaches are predominantly English-focused, and adapting them to other languages is hindered by the scarcity and cost of human-annotated judgments in most languages. We introduce a decomposition-based evaluation framework built around a Universal Criteria Set (UCS). UCS consists of a shared, language-agnostic set of evaluation dimensions, producing an interpretable intermediate representation that supports cross-lingual transfer with minimal supervision. Experiments on multiple faithfulness tasks across languages and model backbones demonstrate consistent improvements over strong baselines without requiring target-language annotations.
\end{abstract}
\section{Introduction}

The rapid growth of general-purpose AI systems has led to a dramatic increase in machine-generated text across applications such as summarization~\cite{pu2023summarization}, question answering~\cite{yue2025survey}, content moderation~\cite{kolla2024llm}, and search~\cite{spatharioti2023comparing}. Assessing the quality and correctness of these outputs at scale remains a central challenge~\cite{wu2025survey, ohde2025burden}. Although human annotation is the gold standard, it is expensive, time-consuming, and difficult to scale to the volume and frequency required by modern development cycles~\cite{ohde2025burden, gu2024survey}.

Traditional automatic evaluation metrics for text generation, such as ROUGE~\cite{lin2004rouge} and BLEU~\cite{papineni2002bleu}, have long served as proxies for output quality. However, they rely primarily on surface-level lexical overlap and fail to capture semantic meaning, reasoning, and contextual appropriateness~\cite{sulem2018bleu}. LLM-based judges can be a more flexible alternative by leveraging broad pretrained knowledge and contextual reasoning. An LLM can be prompted to assess the quality of output using task-specific instructions or rubric-style criteria~\cite{li2025generation, zheng2023judging, chen2024mllm, chiang2024chatbot}, leading to their growing adoption for summarization, factuality assessment, preference modeling, and model comparison.

The development and benchmarking of LLM judges have mainly focused on English, despite only about 15\% of the global population speaking English~\cite{pombal2025m}. As AI systems are deployed worldwide, reliable evaluation across diverse languages becomes a critical bottleneck~\cite{wang2024all, fu2025reliable}. Existing multilingual judge approaches face two primary challenges. First, many rely on language-specific fine-tuning, which requires human-annotated data that is scarce for most languages, and often underperform strong English-based baselines~\cite {doddapaneni2025cross}. Second, English-trained judges do not consistently transfer to typologically distant or low-resource languages~\cite{fu2025reliable}. Together, these limitations hinder scalable and reliable multilingual evaluation. 

These limitations become particularly apparent in the common scenario of language expansion, where systems developed and evaluated in English are later deployed to additional languages. While model capabilities can often be extended through translation or multilingual prompting, evaluation typically requires collecting new annotation data or adapting evaluation frameworks for each language. As a result, evaluation becomes a bottleneck in scaling LLM systems to new linguistic settings, highlighting the need for approaches that can transfer evaluation behavior from English to other languages with minimal additional supervision.

In this paper, we introduce an interpretable decomposition-based framework for LLM judges that enables sample-efficient cross-lingual transfer. Our approach decomposes judgment into a shared set of language-agnostic evaluation criteria, each expressed as a targeted question about a specific dimension of quality. The resulting criterion-level responses define a structured intermediate representation that maps judgments into a predefined, human-interpretable evaluation space, in the spirit of concept-based models that ground intermediate representations in explicit semantic dimensions \cite{koh2020concept, espinosa2022concept, poeta2023concept, sheth2023auxiliary, sunconcept}, while also capturing reasoning patterns that transfer across languages. A lightweight transfer module trained only on English-labeled data can then be applied directly to new languages, enabling cross-lingual judge transfer without target-language supervision.

We evaluate our framework across multilingual benchmarks and multiple LLM backbones. Our contributions are: (1) we show that LLM-judge reasoning can be factored through a shared, language-agnostic criteria space, enabling consistent judgments across heterogeneous linguistic inputs; (2) we introduce an interpretable intermediate representation derived from criteria-level responses; (3) we demonstrate that this representation supports effective cross-lingual transfer using a lightweight module trained in English and applied to other languages without requiring target-language labels.

\section{Related Works}

\paragraph{LLM Judges}
Large language models are increasingly used as automated evaluators for summarization, dialogue, factuality, and preference modeling. Early work \citep{zheng2023judging,liusie2024llm} demonstrated that LLMs can approximate human preferences using carefully designed prompts. Subsequent studies explored rubric-based scoring~\cite{song2024finesure}, pairwise comparison~\cite{liusie2024llm}, and direct answer classification using models like GPT-4. However, these methods typically rely on single-prompt formulations that are brittle to prompt phrasing and struggle to generalize~\cite{thakur2025judging}. 

Another line of work improves evaluation by incorporating explicit reasoning, such as chain-of-thought prompting~\cite{zheng2023judging}, justification-first scoring~\cite{trivediself}, or predefined criteria prompting~\cite{weirocketeval, lee2025checkeval}.

More recent approaches employ multiple LLMs interacting through debate~\cite{feng2025m, chanchateval}, critique~\cite{kim2024debate}, or competitive assessment. While effective in certain settings, these systems require careful human input for selecting few-shot or seed prompts~\cite{feng2025m, lee2025checkeval} and substantial role engineering~\cite{alfano2025multilingual}, making them costly and difficult to deploy reliably in production environments. 

Checklist-style evaluation has recently gained attention as an interpretable alternative to monolithic rubric prompts. Previous work has shown that explicitly decomposing an evaluation task into smaller criteria can improve transparency. \citet{weirocketeval} generate binary checklist items using a stronger model and apply them to smaller evaluators, effectively distilling high-level judgements into simple, verifiable units. \cite{lee2025checkeval} prompt models to generate their own criteria and use these checklists for iterative self-improvement, allowing the evaluator to refine its reasoning over multiple rounds.

\paragraph{Multilingual LLM-Based Judges}
Most LLM-judge design and evaluation has focused on English. \citet{chang2025exploring} studies the impact of resource availability on multilingual evaluators, while \citet{fu2025reliable} analyzes the reliability of multilingual judges. Recent approaches train multilingual evaluators through large-scale pretraining~\cite{pombal2025m} or language-specific fine-tuning~\cite{doddapaneni2025cross}. To reduce reliance on human annotations, \citet{alfano2025multilingual} generates synthetic multilingual supervision by translating English data and constructing corrupted summaries for training, and further examines how training language affects multilingual evaluation behavior; notably, their strongest results are obtained by fine-tuning on English data only.  

However, all of these approaches require either task-specific data construction, full model fine-tuning, or language-specific adaptation. In contrast, our method produces a language-agnostic interpretable representation from a shared criterion set, enabling cross-lingual transfer via a lightweight module trained on only a few labeled English examples with no target-language supervision, no data engineering, and no full-model fine-tuning.

\section{Methodology}

We propose a criteria-based evaluation framework that decomposes an LLM-based judge’s decision into a structured set of sub-criteria. Rather than relying on a single prompt or free-form reasoning, our judge produces answers to a set of targeted evaluation questions, which we refer to as criteria, and these answers are aggregated into an intermediate judge representation used for cross-lingual transfer. 

\paragraph{Problem Formulation.} Consider an evaluation task in which an input sample $x$ must be assigned a binary label $y \in \{0,1\}$. Let $\mathcal{D} = \{(x_i, y_i)_{i=1}^{N}\}$ denote a labeled dataset, where $y_i$ is the ground-truth judgment for sample $x_i$. Each sample $x$ can consist of any structured input relevant to the task, such as a source–output pair, an instruction–response pair, or any text bundle under evaluation. A standard LLM-based judge is obtained by prompting the model $\mathcal{M}$ with an evaluation instruction to produce a predicted label $\hat{y}_i$: \begin{equation} \hat{y}_i = \mathcal{M}(x_i). \end{equation} Standard prompting produces $\hat{y}_i$ directly, either with chain-of-thought reasoning or via multi-agent orchestration.

In this work, we replace this single-prompt-based judgment with a decomposition-based framework in which the LLM first responds to a set of evaluation criteria. These responses form an intermediate judge representation that is used for transfer across languages.

\begin{figure*}[htb!]
\centering
    \includegraphics[width=0.8\textwidth]{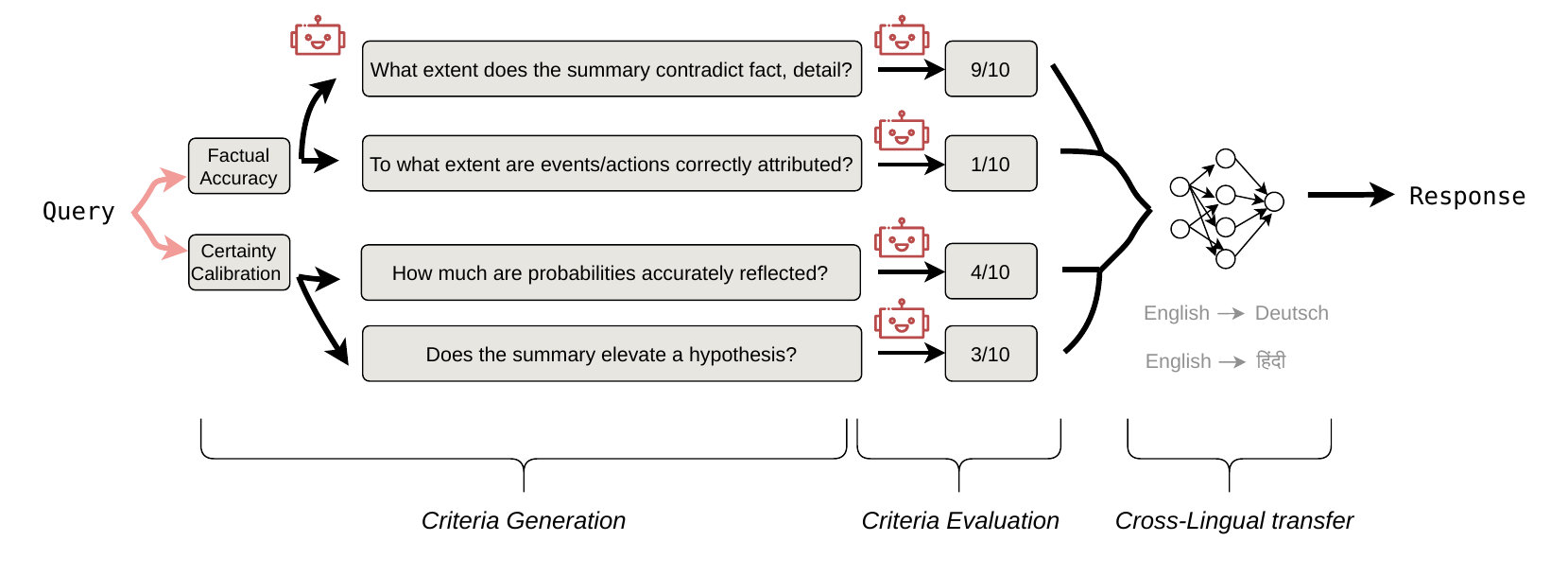}
\caption{UCS framework. The LLM evaluates an input using criteria questions derived from higher-level concepts. Criterion scores are aggregated into a latent representation, which a transfer module trained on English-labeled data maps to the final judgment and applies across languages.}
    \label{fig:ucs_framework}
\end{figure*}

\paragraph{Framework Overview.} Our methodology consists of three stages. \textbf{Stage 1}: \textit{Criteria Set Generation} constructs a set of language-agnostic evaluation criteria from the task specification. \textbf{Stage 2}: \textit{Criteria Set Evaluation} 
% \zeno{Criteria-Set Evaluation?} 
applies these criteria to each input sample, producing a structured intermediate representation from the LLM's criterion-level responses. \textbf{Stage 3}: \textit{Cross-Lingual transfer} trains a lightweight transfer module on labeled English data to align the judge representation, enabling transfer to other languages. An overview of the framework is shown in ~\autoref{fig:ucs_framework}.

\subsection{Criteria Set Generation}

Our approach first constructs a set of evaluation criteria that define the dimensions along which the LLM-based judge evaluates each sample. Formally, we generate a \emph{Universal Criteria Set} (UCS), a collection of evaluation questions
\begin{equation}
    Q = \{ q_1, \ldots, q_k \}
\end{equation}
where each $q_k$ specifies a distinct judgment dimension. We generate criteria in English, motivated by evidence that LLMs exhibit more consistent reasoning and intermediate representations in English than in other languages \cite{schut2025multilingual, shilanguage, huang2024mindmerger}. This choice simplifies the generation of criteria and improves stability for downstream cross-lingual transfer.

The UCS defines a reusable set of evaluation dimensions applicable to all samples for a given task and is shared across all languages. Given a task description $t$, the LLM generates a set of evaluation concepts $C = \{ c_1, \ldots, c_m \}$ representing generic evaluation dimensions such as accuracy, consistency, attribution, or hallucination. The UCS is task-specific, generated from the task description but \textit{language-universal}: the same criteria set is applied to all languages for a given task.

\begin{equation}
C = \mathcal{M}(t)
\end{equation}
See Appendix~\ref{app:prompts} for the prompts used in this stage.

\paragraph{Question Generation} 
For each concept $c_j \in C$, the LLM generates one or more evaluation questions to comprehensively cover its sub-dimensions. Let $Q_j$ denote the subset of criteria generated from the concept $c_j$, such that :
\begin{equation}
    Q_j = \mathcal{M}(c_j)
\end{equation}
The final Universal Criteria Set is obtained by aggregating the criteria generated for all concepts:
\begin{equation}
        Q = \bigcup_{j=1}^{m} Q_j.
\end{equation}
Thus, $Q = \{ q_1, \ldots, q_k \}$ represents the union of all concept-derived questions. The same criteria set $Q$ is applied to every sample, providing stable language-agnostic evaluation features that support cross-lingual transfer.

\subsection{Criteria Set Evaluation}
% \zeno{Criteria-Set Evaluation?}
In the second stage of our framework, given the UCS $Q$ and an input sample $x$, the LLM is prompted to assess the extent to which $x$ satisfies each criterion $q_k \in Q$, returning a numerical score:
% \zeno{PLLM}
\begin{equation}
\mathcal{M}(q_k, x)= z_k(x) \in [1, \ldots, 10],
\end{equation}
We consider Likert rating from 1-10 following \cite{lee2025checkeval}. Other rating ranges is left to future explorations. Collecting these outputs yields the \emph{criteria-response vector}: 
\begin{equation}
    z(x) = [\, z_1(x), \ldots, z_k(x) \,],
\end{equation}
which serves as a structured and interpretable intermediate judge representation of the sample.

\subsection{Cross-Lingual Transfer}

The final stage learns a mapping from the criteria-response representation $z(x)$ to a calibrated judgment that generalizes across languages. The key insight is that while raw LLM outputs may vary across languages, the structure of evaluation as expressed through the shared criteria remains stable. The criteria-response vector $z(x)$ therefore defines a language-agnostic intermediate representation space in which a lightweight predictor trained on one language can be applied directly to another.

\paragraph{Concept-Level Aggregation.}
Rather than operating on the full criteria-response vector $z(x)$, we perform concept-level transfer by aggregating criterion-level scores within each concept. This reduces the dimensionality of the representation, decreases sample complexity, and improves transfer robustness (see Appendix~\ref{app:conceptlevel} for details). Concretely, for each concept $c_j$, we average the scores of its associated criteria:
\begin{equation}
    s_j(x) = \frac{1}{|Q_j|} \sum_{q_i \in Q_j} z_i(x).
\end{equation}
The resulting concept-level representation is
\begin{equation}
    s(x) = [\, s_1(x), \ldots, s_m(x) \,].
\end{equation}

\paragraph{Transfer.}
Given labeled data in a source language $\ell$, we train a lightweight predictor (e.g. a neural network) on the concept-level representation:
\begin{equation}
    \hat{y} = f_\theta(s(x)),
\end{equation}
where $\theta$ is learned on the source-language dataset $\mathcal{D}^\ell = \{(x_i^\ell, y_i^\ell)\}_{i=1}^N$. At inference time, the trained predictor $f_{\theta^\ast}$ is applied directly to samples in a target language $\ell'$:
\begin{equation}
    \hat{y}_j^{\ell'} = f_{\theta^\ast}(s(x_j^{\ell'})).
\end{equation}
This enables cross-lingual transfer without any target-language supervision, as the shared criteria space ensures that $s(x)$ remains semantically consistent across languages.

\begin{table*}[htb!]
\small
\centering
\begin{tabular}{lcccccccccccccc}
\toprule
Method 
& \multicolumn{5}{c}{MEMERAG} 
& \multicolumn{9}{c}{mFACE} \\
\cmidrule(lr){2-6} \cmidrule(lr){7-15}
& DE & FR & ES & HI  & \textit{Avg}
& AM & MY & FR & SW & TH & AR & HI & ES & \textit{Avg} \\
\midrule

Zero-shot  & 77.2 & 77.5 & 77.5 & 74.4 & 76.7 & 54.7 & \textbf{70.2} & 74.3 & 73.0 & 72.6 & 69.7 & 68.0 & 77.9 & 70.1 \\
CoT        & 73.0 & 76.2 & 76.2 & 74.1 & 74.9 & 58.7 & 68.5 & 72.8 & 77.5 & 71.4 & 69.2 & 69.4 & 74.4 & 70.2 \\
AG         & 77.2 & 80.6 & 76.1 & 78.0 & 78.0 & 59.3 & 67.4 & 76.6 & 77.7 & 72.6 & 70.7 & 69.0 & 79.5 & 71.6 \\
ChatEval   & 71.8 & 73.0 & 60.9 & 62.4 & 67.0 & 51.1 & 65.5 & 68.4 & 74.0 & 63.9 & 66.1 & 55.7 & 70.3 & 64.4 \\
CheckEval  & 76.1 & 80.7 & 74.8 & 78.4 & 77.5 & 60.8 & 68.9 & 74.9 & 76.8 & \textbf{75.1} & 73.4 & 65.8 & 76.4 & 71.5 \\
RocketEval & 76.4 & 81.0 & 75.9 & 79.1 & 78.1 & 61.2 & 69.2 & 75.4 & 77.0 & 74.9 & 72.9 & 66.9 & 78.1 & 72.0 \\
\rowcolor{lightblue}
UCS (EN)   & \textbf{79.7} & \textbf{84.3} & \textbf{80.0} & \textbf{82.2} & \textbf{81.6} & \textbf{64.7} & 67.7 & \textbf{82.4} & \textbf{79.0} & \textbf{75.1} & \textbf{74.6} & \textbf{70.0} & \textbf{85.5} & \textbf{74.9} \\
\bottomrule
\end{tabular}
\caption{Balanced Accuracy (BA) across methods and languages for the Qwen-235B model. Bold values indicate the best performance per column averaged across 3 runs. Our method UCS is trained on English only data (EN).}
\label{tab:T1}
\end{table*}

\section{Experimental Setup}

\paragraph{Models.} We evaluate our methods across a range of LLMs to ensure diversity in scale and architecture. We report the results for Qwen3-32B~\cite{yang2025qwen3}, Qwen3-235B-A22B~\cite{yang2025qwen3}, OSS-20B~\cite{agarwal2025gpt},  and OSS-120B~\cite{agarwal2025gpt}. All LLM-based judges are queried in a deterministic setting with \texttt{temperature}~=~0 and \texttt{top\_p}~=~1.0 to reduce randomness in evaluation outputs.

\paragraph{Datasets.} Our primary experiments use two multilingual evaluation benchmarks. \textbf{MEMERAG}~\cite{blandon2025memerag} is a multilingual RAG faithfulness benchmark containing evidence--query--answer triples, each annotated with a binary faithfulness label. The dataset spans five languages: English (EN), French (FR), German (DE), Hindi (HI), and Spanish (ES). \textbf{mFACE}~\cite{aharoni2023multilingual} is a multilingual summarization evaluation dataset consisting of news articles and human-written summaries collected from BBC regional editions. The task is to judge whether a summary is faithful to the source article. We use a representative subset of languages spanning high-resource, mid-resource, and low-resource languages: English (EN), Amharic (AM), Burmese (MY), French (FR), Swahili (SW), Thai (TH), Arabic (AR), Hindi (HI), and Spanish (ES). 
% not MT, different levels of difficulty
\paragraph{Evaluation Metrics.}
We report \emph{balanced accuracy} (BA) to account for class imbalance, as it measures judge's sensitivity to both positive and negative classes equally.

\paragraph{Implementation Details.}
For each dataset and model, all methods are evaluated using consistent prompting templates and fixed criteria-generation procedures. All transfer models are trained exclusively on English-labeled data unless otherwise noted, and are applied to the remaining languages without any target-language supervision. We use shallow neural network for transfer from English to other languages (See ~\ref{app:hyperparam} for details).
% add ref to app

\paragraph{Baselines.}
We evaluate our approach against a broad set of strong LLM-based judge baselines spanning single-model prompting, multi-agent methods, and checklist-based evaluators. First, we include zero-shot LLM judges following standard prompting setups from previous work \cite{blandon2025memerag, bavaresco2025llms}, where the model is directly instructed to assess correctness or faithfulness without additional structure. We also include chain-of-thought (CoT) prompting, which has been shown to improve reasoning in LLM judges~\cite{zheng2023judging}.
% \zeno{Next 2 sentences are not clear: what are annotation guidelines? citation? how is that used? why is not realistic? } 
Another baseline we consider is prompting LLM with annotation guidelines (AG) that were given to humans, similar to ~\cite{blandon2025memerag}, see Appendix \ref{app:reprod}. 

We further compare against ChatEval~\citep{chanchateval}, a multi-agent debate-style evaluator in which models critique and challenge each other's assessments. Such systems have demonstrated strong performance on complex reasoning and judgment tasks, but require substantial orchestration overhead. Finally, we include two recent checklist-based evaluation frameworks: RocketEval~\cite{weirocketeval} and CheckEval~\cite{lee2025checkeval}, which generate or refine structured criteria to guide LLM judgments. These methods share our goal of improving structure and interpretability in LLM-based evaluation; however, unlike our approach, they do not produce a unified latent judge representation that supports cross-lingual transfer~\footnote{We reproduce the baseline results using publicly code.}.

\section{Results}
In this section, we evaluate the proposed criteria-based framework across languages and datasets. We focus on the language expansion setting and report the performance of UCS when trained on the English portion of the multilingual dataset, denoted by UCS (EN). We then present a series of analyses that unpack the sources of these gains: sample-efficiency curves quantifying how much English supervision is required to reach stable cross-lingual performance (§~\ref{sec:sample_efficiency}), an analysis studying the most predictive criterion dimensions (§~\ref{sec:criteria_importance}), a comparison of alternative transfer models (§~\ref{sec:alternative_models}), a comparison of criterion aggregation strategies (§~\ref{sec:aggregation}) and an analysis of inference cost trade-offs (§~\ref{sec:inference_cost}).

\subsection{Cross-Lingual Transfer}

% \zeno{Ivaxi - I have completely rewritten this subsection, pls review.}
~\autoref{tab:T1} reports balanced accuracy (BA) for the Qwen-235B judge on MEMERAG and mFACE, comparing the proposed UCS framework trained on English data against prompting-based, debate-style, and checklist-based baselines. Across both datasets, UCS (EN) delivers the strongest and most consistent cross-lingual performance, achieving the best or tied-best results in the majority of evaluated languages (11 out of 12). Appendix Tables~\ref{tab:T1_3run_diverse_memerag} and~\ref{tab:T1_3run_diverse_mFACE} report results averaged over multiple random seeds, showing that UCS maintains consistent improvements across languages while exhibiting low variance across runs.

On MEMERAG, UCS (EN) achieves the highest performance across all four non-English languages, with an average BA of 81.6 compared to 78.1 for RocketEval, the strongest baseline.   Improvements are consistent across typologically diverse languages, ranging from 3.1 to 4.1 BA points and spanning Germanic (DE), Romance (FR, ES), and Indo-Aryan (HI) languages. This uniformity supports the claim that the criteria-based representation is genuinely language-agnostic for this task.

On mFACE, UCS (EN) achieves best-or-tied performance on seven of the eight evaluated languages, with an average BA of 74.9 compared to 72.0 for RocketEval. Notably, the largest improvements are observed for French (+7.0) and Spanish (+7.4) -- two high-resource languages where strong baseline performance might be expected. This suggests that the structured criteria space captures evaluation dimensions relevant to summarization faithfulness that are not easily elicited through direct prompting, and that this benefit is most pronounced for languages where the LLM can reliably interpret and respond to the criteria. Consistent gains are also observed on lower-resource languages such as Amharic (+3.5) and Hindi (+3.1), further demonstrating the breadth of cross-lingual generalization. 

One notable exception is observed on Burmese (MY), where the zero-shot baseline achieves the highest performance across all methods (70.2), outperforming not only UCS (67.7) but also all other baselines. This suggests that Burmese may exhibit language-specific properties, such as script complexity or limited LLM pretraining coverage, that are not well captured by the shared criteria space, and that direct prompting may be more robust in such cases. Despite this exception, UCS maintains the strongest overall performance profile across languages, demonstrating the robustness of the proposed criteria-based representation for cross-lingual judge transfer.Appendix ~\autoref{app:fullcrosslingual} reports the full cross-lingual results across all judge backbones evaluated in this work. The results are consistent with the main findings: UCS-based judges achieve the best or near-best performance across most language–model combinations on both MEMERAG and mFACE, indicating that the improvements are not specific to an LLM backbone.

\subsection{Sample Efficiency}
\label{sec:sample_efficiency}

A practical consideration for language expansion is how much labeled data in a high-resource language is required to enable effective transfer. Although our framework trains only a lightweight transfer module on English-labeled criterion features, understanding the supervision requirements for stable cross-lingual performance is important for assessing the practical viability of the approach.

To study this, we construct a fixed random split of the English data, holding out 30\% as a test set. From the remaining 70\%, we vary the amount of labeled training data used to fit the transfer module, ranging from 5\% to 100\% of the available training portion. Each resulting model is evaluated both on the held-out English test set and on all target languages, allowing us to examine how increasing English supervision affects both in-language performance and cross-lingual generalization.

\begin{figure}[t!]
\centering
\begin{subfigure}[b]{0.48\linewidth}
  \centering
  \includegraphics[width=\linewidth]{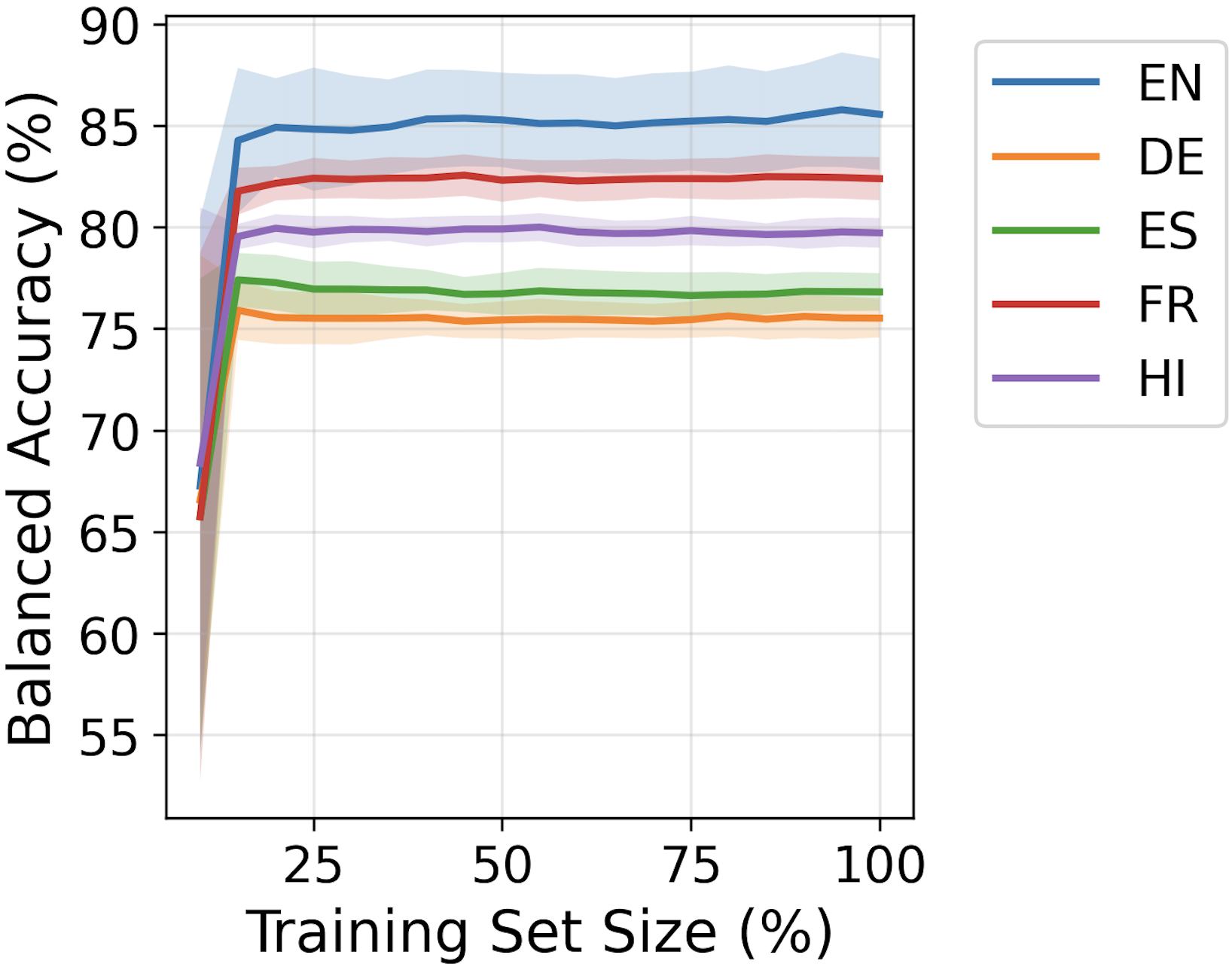}
  \caption{}
\end{subfigure}
\hfill
\begin{subfigure}[b]{0.48\linewidth}
  \centering
  \includegraphics[width=\linewidth]{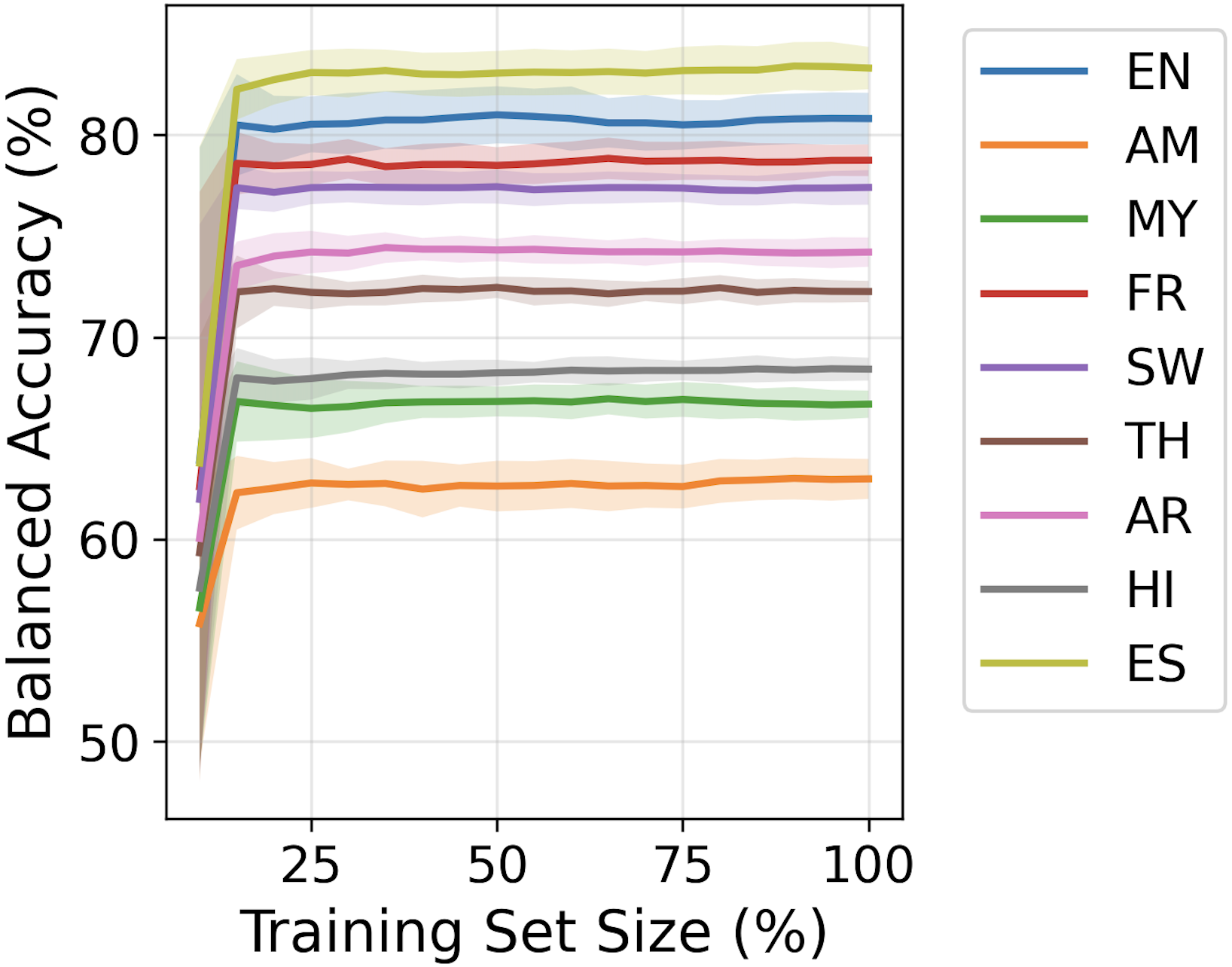}
  \caption{}
\end{subfigure}
\caption{Sample efficiency of training the transfer module on English data. Performance is shown as a function of the percentage of English training data used. Results are reported for MEMERAG (a) and mFACE (b).}
\label{fig:sample_Eff}
\end{figure}

~\autoref{fig:sample_Eff} shows that performance improves rapidly with a small fraction of labeled data and stabilizes once approximately 20--30\% of the English training data is used. Beyond this point, additional supervision yields only marginal gains across all languages.

Importantly, the same trend is observed for both English and target languages: as the transfer module improves on English, performance increases consistently across languages. This suggests that the criteria-based intermediate representation enables efficient cross-lingual generalization, requiring only a modest amount of labeled data in a single high-resource language.

\subsection{Criteria Importance}
\label{sec:criteria_importance}

Our transfer framework assumes that the criteria representation captures evaluation signals that are meaningful across languages. A natural question is whether the relative importance of these evaluation dimensions is preserved across languages. We hypothesize that languages whose importance profile criteria are more similar to English should exhibit stronger transfer performance.

To examine this, we analyze the importance of individual criteria before concept aggregation. While the transfer model operates on concept-level averages, criterion-level analysis provides a finer diagnostic view of how evaluation signals are prioritized across languages.

\begin{figure}[t!]
\centering
\begin{subfigure}[b]{0.48\linewidth}
  \centering
  \includegraphics[width=\linewidth]{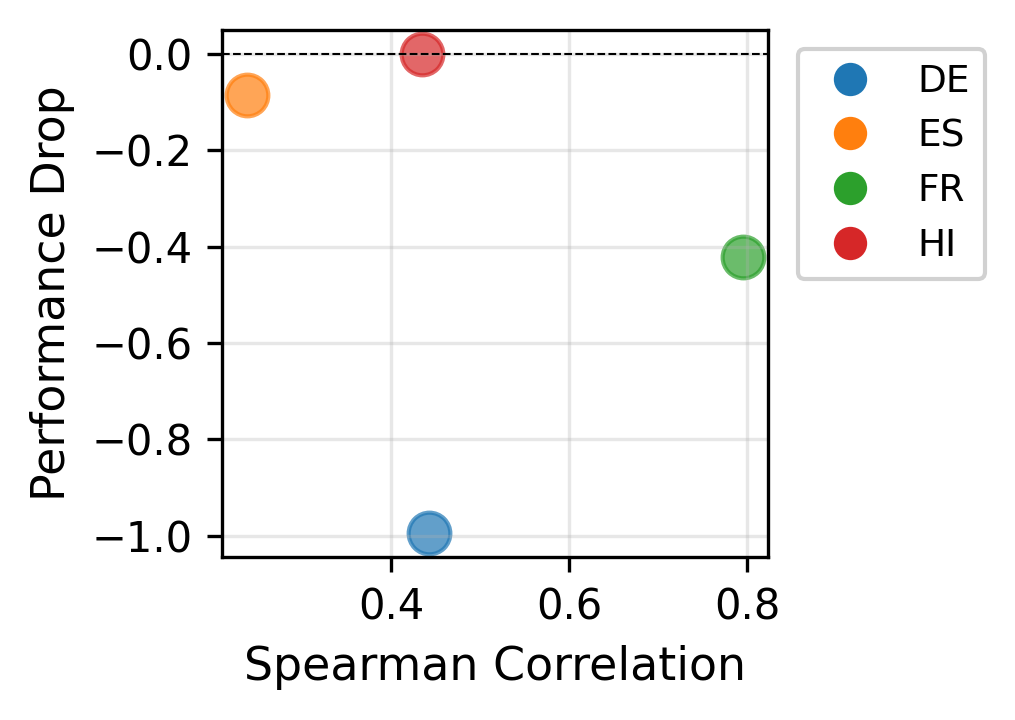}
  \caption{MEMERAG}
\end{subfigure}
\hfill
\begin{subfigure}[b]{0.48\linewidth}
  \centering
  \includegraphics[width=\linewidth]{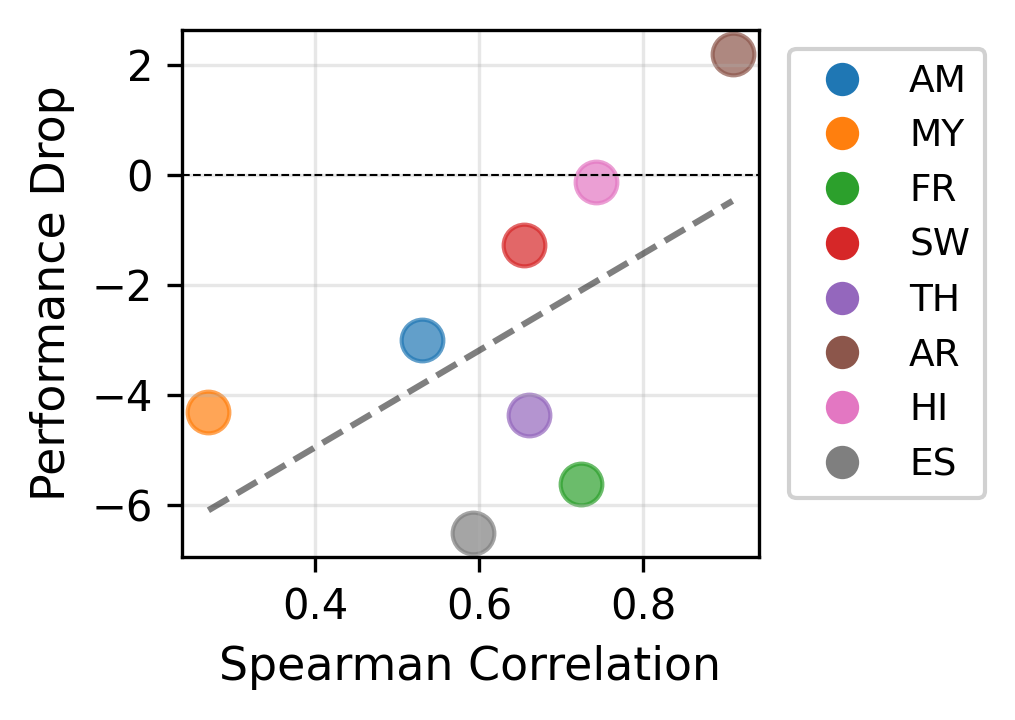}
  \caption{mFACE}
\end{subfigure}
\caption{Relationship between English–target criteria importance alignment and performance change when training with only the top-10 English-selected criteria. 
}
\label{fig:correlationvsperf}
\end{figure}

For each language $\ell$, we train a Random Forest classifier using that language’s criterion responses together with the corresponding human labels, and extract feature importance scores using Gini impurity reduction~\cite{nembrini2018revival}. We then compute the Spearman rank correlation~\cite{spearman1961proof} between the English importance profile and the language-specific importance profile. To evaluate whether this alignment matters in practice, we simulate a constrained setting where only the top-$10$ criteria selected according to English importance are used to train the classifier for each language. We then measure the performance change relative to using the full criteria set. ~\autoref{fig:correlationvsperf} plots, for each language, the relationship between English–target importance correlation and the resulting performance change.

In mFACE, we observe that languages with lower importance alignment generally experience larger performance degradation. For example, Burmese (MY) shows one of the lowest correlations with English and also corresponds to one of the weaker-performing languages in ~\autoref{tab:T1}. While this observation is only suggestive, it indicates that differences in evaluation priorities across languages can contribute to variation in cross-lingual transfer performance. For MEMERAG, we observe only minor performance differences when restricting the model to English-selected criteria; the drops remain below 1\% across languages and do not show a clear relationship with importance alignment.

~\autoref{fig:correlation} visualizes full distribution of importance scores across all evaluated languages where some languages exhibit importance profiles similar to English, others show noticeable differences in how evaluation dimensions are prioritized. 
Overall, these results indicate that the relative importance of evaluation criteria is not universally shared across languages. Selecting a fixed subset of criteria based solely on English importance can harm performance for languages with different importance profiles. This observation supports the design of our full criteria-based representation, which allows the transfer module to learn language-appropriate weightings rather.

\subsection{Alternative Training Models}
\label{sec:alternative_models}

While our primary transfer model uses a shallow neural network to map criteria-response representations to final judgments, we also evaluate alternative transfer modules to assess the sensitivity of our framework to the choice of predictor. These include logistic regression as a linear baseline, SVMs for margin-based classification, KNN as an instance-based method, and tree-based models such as Random Forests and XGBoost that can capture nonlinear feature interactions.

\begin{table}[t!]
\footnotesize
\centering
\begingroup
\begin{tabular}{>{\columncolor{white}}l|cc}
\toprule
Method & MEMERAG & mFACE \\
\midrule
LogReg  & 80.9 & 78.2 \\
SVM     & 78.5 & \textbf{79.0} \\
KNN     & 77.1 & 70.9 \\
RF      & 76.4 & 69.8 \\
XGBoost & 74.3 & 71.1 \\
NN      & \textbf{81.5} & 78.9 \\
\bottomrule
\end{tabular}
\endgroup
\caption{BA averaged across languages for different transfer modules on the Qwen-235B model.}
\label{tab:trainingmodules}
\end{table}

~\autoref{tab:trainingmodules} shows that several simple models perform competitively when operating on the criteria-based representation. Logistic regression and SVM achieve strong results, indicating that much of the signal captured by the criteria representation is linearly separable.

The shallow neural network achieves the best performance on MEMERAG and remains competitive on mFACE, where SVM slightly outperforms it. Overall, the neural network provides the most consistent performance across datasets and languages.

In contrast, instance-based and tree-based methods perform substantially worse. These models appear to overfit to patterns in the English training data and generalize less effectively to other languages, suggesting that simpler parametric models are better suited for cross-lingual transfer here.

\subsection{Transfer Learning vs LLM-based Aggregator}
\label{sec:aggregation}
The criteria-response vector $z(x)$ produced by the second stage must be aggregated into a final binary judgment. In our primary framework, this is achieved by the lightweight transfer module $f_\theta$ trained on English-labeled data, which learns a calibrated mapping from the concept-level representation $s(x)$ to the predicted label $\hat{y}$.

An alternative we consider an \emph{LLM-based aggregator}, in which the LLM is prompted to produce a final judgment given the input $x$, the criteria $Q$, and their corresponding responses $z(x)$. This approach requires no additional training and may leverage the LLM's broader world knowledge and reasoning capabilities to capture interactions between criteria. However, it does not benefit from calibration on human-labeled data and may be sensitive to prompt phrasing and model-specific biases.

\begin{table}[t!]
\small
\centering
\begin{tabular}{l|cc}
\toprule
Method & MEMERAG & mFACE \\
\midrule
UCS (Trained)  & \textbf{81.5} & \textbf{74.9} \\
UCS (LLM) & 77.2 & 74.3 \\
\bottomrule
\end{tabular}
\caption{Average BA across languages on MEMERAG and mFACE. UCS (Trained) uses a lightweight transfer module trained on English-labeled data; UCS (LLM) uses the LLM itself as a training-free aggregator over the criteria responses.}
\label{tab:avg_results}
\end{table}

~\autoref{tab:avg_results} demonstrates a trade-off between training-free and trained aggregation. The LLM-based aggregator does not require labeled data beyond the criteria responses and can be applied directly. However, when labeled English data is available, training a lightweight transfer module on the criteria-based representation consistently yields a higher average balanced accuracy on both datasets.

\subsection{Inference Cost}
\label{sec:inference_cost}
% \zeno{How does it compare against best baseline RocketEval? Or other methods?}
Compared to zero-shot prompting, our criteria-based framework incurs additional inference cost due to explicit criteria generation and criterion-level evaluation. While a zero-shot judge requires a single LLM call per sample, our approach introduces structured evaluation steps that increase prompt length and the number of LLM calls.

To analyze the trade-offs between decomposition granularity and inference efficiency, we evaluate four prompt variants differing in how criteria are generated and scored. Generation can be joint i.e. all evaluation dimensions produced in one LLM call or per-concept, i.e. each concept generates its criteria separately). Scoring follows the same options: a single joint judgment over all criteria or separate per-concept scores.

\begin{table}[t!]
\small
\centering
\begin{tabular}{ll|cc}
\toprule
Generation & Scoring & MEMERAG & mFACE \\
\midrule
Joint       & Joint       & 79.1 & 77.4 \\
Joint       & Per-concept & 81.0 & 78.6 \\
Per-concept & Joint       & 80.2 & 78.0 \\
Per-concept & Per-concept & \textbf{81.5} & \textbf{78.9} \\
\bottomrule
\end{tabular}
\caption{Average BA across languages on MEMERAG and mFACE for UCS under different decompositions. Joint denotes evaluating all criteria in one LLM call, while Per-concept denotes one call per concept.}
\label{tab:cost}
\end{table}

~\autoref{tab:cost} compares these prompt variants in terms of both inference cost and performance. Performance improves consistently as the evaluation process is more finely decomposed. In particular, scoring at the per-concept level consistently outperforms joint scoring over all criteria. The best results are obtained when both criteria generation and scoring are performed at the per-concept level, indicating that finer-grained criterion-level signals provide a more discriminative intermediate representation for the downstream transfer module.

\section{Conclusion}
In this paper, we introduced a decomposition-based framework for LLM judges that enables sample-efficient cross-lingual transfer through a shared set of language-agnostic evaluation criteria. We proposed Universal Criteria Sets (UCS), which structure evaluation into explicit dimensions and produce a transparent intermediate representation of the judgment process. A lightweight transfer module trained on English-labeled data maps this representation to final judgments and generalizes directly to new languages without requiring target-language supervision. Across experiments on MEMERAG and mFACE, UCS achieves strong and consistent performance while requiring only a small amount of supervision in a single high-resource language. Beyond improved cross-lingual performance, the criteria-based representation provides interpretable insights into how evaluation signals contribute to final judgments, and our criteria-importance analysis reveals that the degree of cross-lingual alignment varies by task. Together, these results highlight the potential of structured, criteria-based representations as a principled foundation for building reliable and interpretable multilingual LLM-based judges. The structured nature of this representation also allows future work to explore criterion-level judgments to construct structured reward signals or rubric-based feedback for reinforcement learning and alignment. 

\section*{Limitations}

First, the approach inherits the sensitivity from the underlying LLM. Criterion-level responses may vary with prompt phrasing, decoding settings, or model updates. Although decomposition reduces some instability by structuring evaluation through fixed criteria, the system remains dependent on the robustness of the base model. Future work could investigate prompt-invariant formulations or uncertainty-aware calibration. Second, our method assumes that evaluation criteria capture stable semantic dimensions shared across languages. This assumption may not hold for culturally specific judgments, stylistic norms, or tasks where evaluation standards differ substantially across regions. Finally, the framework relies on English supervision to learn the transfer module. Although this reduces the need for multilingual labels, it can introduce biases present in English evaluation data that need to be studied. Finally, we do not systematically explore the impact of variance that cascades through each stage of UCS generation. 
\section{Future Work}

In this work, we evaluated cross-lingual transfer across multiple languages and datasets. Future work can extend this to other languages not evaluated here. In particular, language-specific analyses could help explain the strong zero-shot performance of the LLM Judge in Burmese. ~\autoref{fig:correlationvsperf} illustrates the relationship between English–target criteria importance alignment, showing that the relative importance of evaluation criteria can vary across languages and datasets. This suggests that not all criteria contribute equally to evaluation quality in cross-lingual settings. In particular, some criteria appear consistently less important across languages, while others show stronger and more stable alignment. These observations indicate the need to identify the most effective evaluation criteria for multilingual assessment. Future work could focus on selecting or learning criteria that provide the strongest signal across languages.

Beyond cross-lingual evaluation, the criteria-based decomposition introduced in this work opens several promising directions. Because the framework represents judgments through explicit evaluation dimensions, it provides a structured interface for analyzing and controlling LLM evaluation behavior. 

Another promising direction is leveraging criteria representations to support human-in-the-loop evaluation. Structured criteria could allow human evaluators to provide targeted feedback on specific dimensions, which can then be incorporated to refine the transfer model or adjust evaluation standards. This may enable more transparent and controllable evaluation pipelines.

Finally, the criteria representation provides a natural foundation for studying the reliability of LLM judges. Future work could investigate uncertainty estimation, agreement across multiple judge models, or ensemble approaches that operate at the criterion level.

Future work could also explore how criterion-level representations can support reinforcement learning and alignment of language models. Current alignment methods often rely on scalar reward signals derived from preference comparisons or holistic judgments, which provide limited insight into why a response is preferred. In contrast, criterion-based evaluation decomposes quality into explicit dimensions, such as factual consistency, relevance, or completeness. These structured signals could be used to construct richer reward functions that guide models toward satisfying multiple evaluation dimensions simultaneously. Moreover, criterion-level feedback may enable more interpretable and controllable alignment, allowing training objectives to emphasize specific aspects of behavior or adapt across languages and domains. Investigating how such structured evaluation signals can be integrated into reinforcement learning or preference optimization pipelines is a promising direction for future work.

\subsection{Risks}

Our framework introduces several potential risks. First, reliance on English-labeled data for training the transfer module may propagate biases present in English evaluation standards, potentially leading to unfair or misaligned judgments in other languages. Second, the assumption of language-agnostic evaluation criteria may overlook culturally specific norms, stylistic preferences, or context-dependent interpretations of quality, resulting in systematic evaluation errors. Third, the approach depends on the stability of underlying LLM outputs; variations due to prompt phrasing or model updates may affect criterion-level responses and downstream predictions. Finally, as with other automated evaluators, there is a risk of over-reliance on LLM-based judgments in high-stakes settings without sufficient human oversight.

% Bibliography entries for the entire Anthology, followed by custom entries
% \bibliography{anthology,custom}
% Custom bibliography entries only
\bibliography{latex/custom}
\newpage
~\newpage
\appendix
\label{sec:appendix}

\section{Reproducibility}
\label{app:reprod}
We will release our code. All criterion-generation and scoring prompts are included in the appendix.

\subsection{Model Inference Settings.}
All LLM-based evaluations were performed with the temperature set to 0 and the top-$p$ set to 1 to ensure deterministic outputs. We report the exact model snapshots used in our experiments. All models were accessed through the Amazon Bedrock API. No additional fine-tuning of the backbone LLMs was performed. We reproduced the results for all of the baselines reported in the paper.

\subsection{Cross-Lingual Transfer Setup.}
For concept-level transfer, we train calibration models on English criterion-level representations and evaluate zero-shot on other languages without using any target-language labels. All experiments use \texttt{random\_state=42} for reproducibility.

\subsection{Transfer Module Hyperparameters.}
\label{app:hyperparam}
We evaluated multiple lightweight predictors.
We evaluated several lightweight calibration models for concept-level transfer. The neural network consists of a single hidden layer with 32 units, trained with a learning rate of 0.01 for up to 2000 iterations. Logistic regression uses a regularization strength of C = 0.1, is trained for up to 2000 iterations, and applies balanced class weights to account for label imbalance. The random forest model uses 200 trees with a maximum depth of 10 and balanced class weights. The support vector machine uses C = 0.1, a gamma value set to “scale” enables probability estimates, and applies balanced class weights. Gradient boosting is configured with 100 estimators, a maximum depth of 5, and a learning rate of 0.1. Finally, the k-nearest neighbors classifier uses k = 5 neighbors.

Unless otherwise stated, the reported results correspond to the NN model selected on English validation data.

\subsection{Data Splits and Evaluation.}
We train the transfer module exclusively on labeled English data and evaluate on multilingual test splits without retraining or hyperparameter tuning. Balanced accuracy is used as the primary evaluation metric to account for label imbalance.

\subsection{Implementation Details.}
All experiments were implemented in Python using standard machine learning libraries. Fixed random seeds were used across training runs. During transfer, no target-language supervision, translation, or synthetic augmentation was used .

\subsection{Annotation Guideline baseline.}
A baseline we consider prompts the LLM with the annotation guidelines (AG) originally provided to human annotators, following the setup of \citet{blandon2025memerag}. For MEMERAG, the annotation guidelines are available in Appendix A of the paper and describe the criteria used by annotators to assess faithfulness between the generated answer and the supporting evidence. For mFACE, we use the evaluation instructions provided in Figure 2 of the paper, which outline the conditions under which a summary should be considered faithful to the source article. In both cases, these guidelines are directly incorporated into the evaluation prompt to guide the LLM’s judgment.

\section{Concept-level transfer}
\label{app:conceptlevel}
A central design choice in our framework is to perform transfer at the level of evaluation concepts rather than at the level of raw criterion interactions. Each dimension in $z(x)$ corresponds to a semantically meaningful evaluation question (e.g., faithfulness, completeness, consistency). By learning a linear calibration over these concept-level signals, the transfer module estimates how much each evaluation dimension contributes to the final judgment.

An alternative would be to model interactions between criteria using a more expressive predictor. However, such approaches substantially increase the number of learnable parameters and, consequently, the number of labeled samples required for stable training. In multilingual settings where supervision is typically available only in English, this would lead to overfitting and poor generalization.

By constraining transfer to operate over independent semantic dimensions, we reduce sample complexity and improve stability. This design aligns with the intuition that evaluation structure is largely shared across languages, even when surface realizations differ. As a result, concept-level calibration enables effective cross-lingual transfer using limited source-language supervision.

\section{Results}\label{app:results}

Table ~\ref{app:fullcrosslingual} reports the full cross-lingual evaluation results for all baselines and criteria-based methods across models and languages on MEMERAG and mFACE. Consistent with the main results, UCS-based judges generally achieve the strongest or among the strongest performance across languages and model backbones.

\begin{figure*}[h]
\centering
    \begin{subfigure}[b]{0.48\linewidth}

         \centering
        \includegraphics[width=\linewidth]{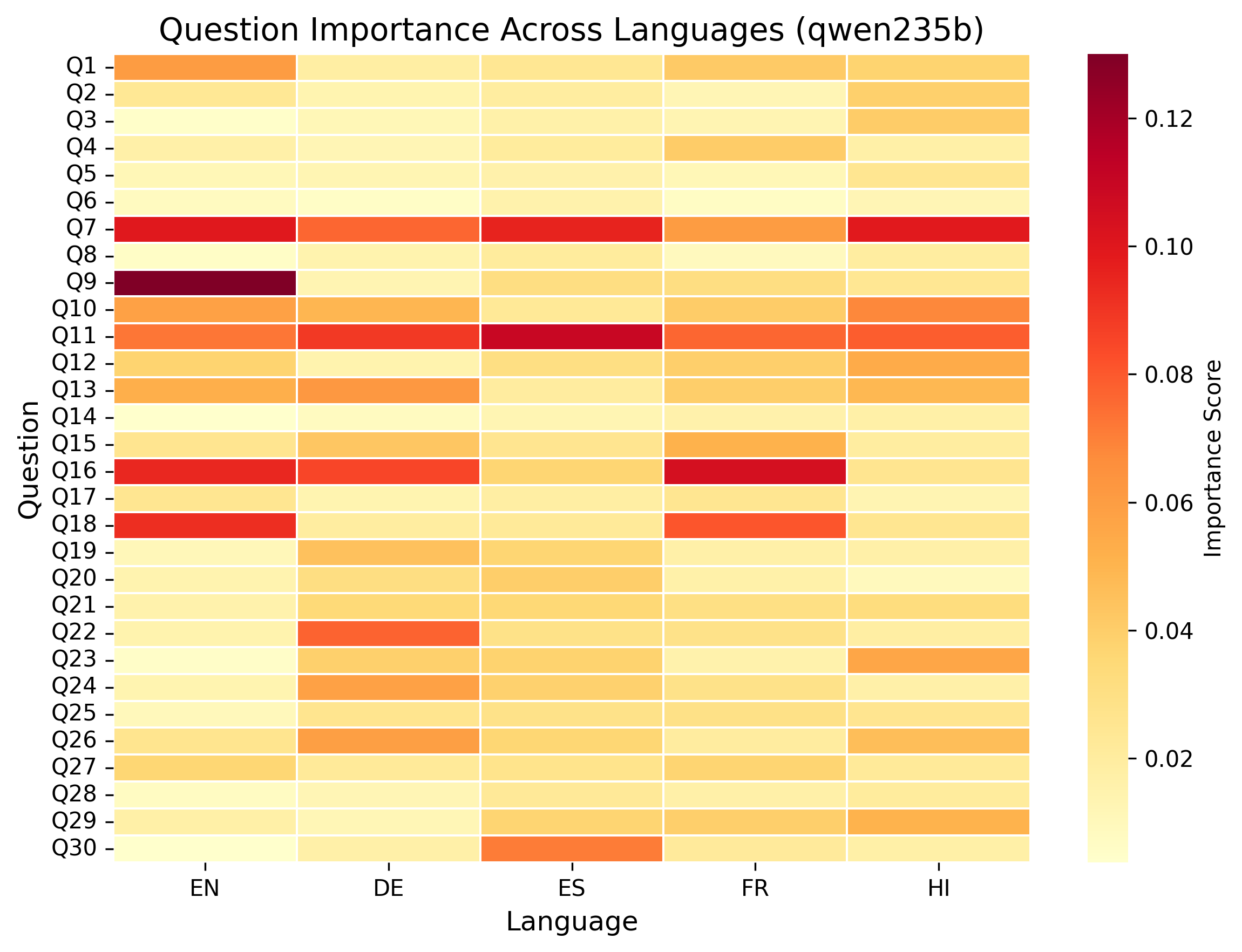}
     \end{subfigure}
     % \hfill
     \begin{subfigure}[b]{0.48\linewidth}

         \centering
         \includegraphics[width=\linewidth]{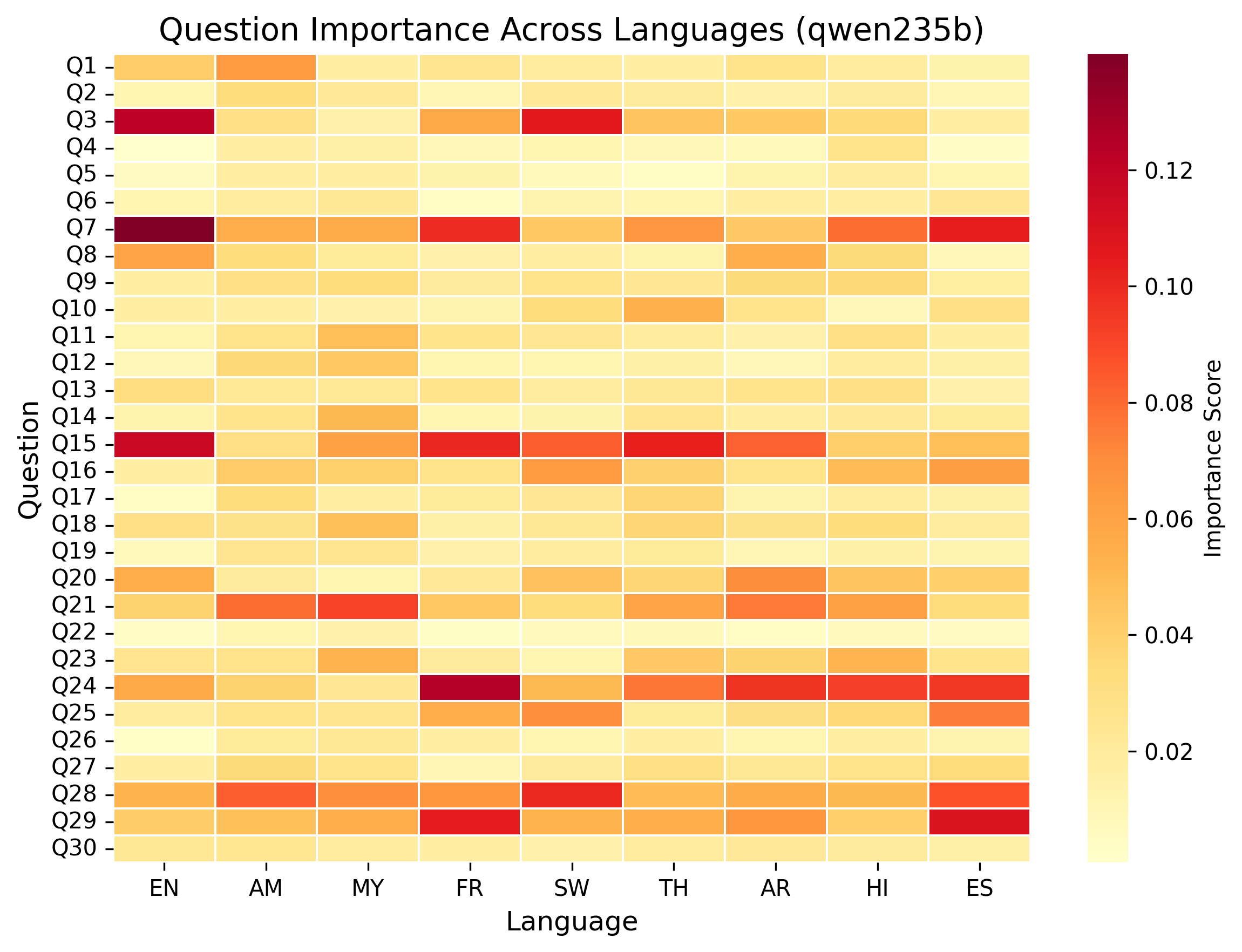} 
     \end{subfigure}
\caption{Relationship between English–target criteria importance alignment. 
} 
\label{fig:correlation}
\end{figure*}

\begin{table*}[h]
\tiny
\centering
\begin{tabular}{llcccccccccccc}
\toprule
\multirow{3}{*}{Model} & \multirow{3}{*}{Method} 
& \multicolumn{4}{c}{MEMERAG (BA)} 
& \multicolumn{8}{c}{mFACE (BA)} \\
\cmidrule(lr){3-6} \cmidrule(lr){7-14}
& & DE & FR & ES & HI 
  & AM & MY & FR & SW & TH & AR & HI & ES \\
\midrule

% ================= Llama-3.3 8B =================

% ================= OSS-20B =================
% ================= OSS-20B =================
\multirow{10}{*}{OSS-20B}
& Zero-shot  & 70.7 & 81.6 & 80.4 & 81.7 & 59.4 & 65.6 & 73.7 & 79.2 & 76.8 & 69.2 & 68.5 & 78.0 \\
& CoT        & 70.5 & 81.6 & 82.0 & 80.6 & 60.9 & 64.9 & 74.6 & 76.2 & 78.3 & 65.2 & 67.6 & 78.0 \\
& AG         & 71.7 & 80.8 & 75.6 & 80.9 & 61.4 & 62.2 & 75.2 & 77.8 & 76.9 & 72.4 & 66.7 & 78.0 \\
& ChatEval   & 64.2 & 73.5 & 71.4 & 72.0 & 54.7 & 60.1 & 66.9 & 70.3 & 68.4 & 62.9 & 61.4 & 71.1 \\
& CheckEval  & 72.5 & 81.9 & 79.9 & 80.8 & 60.9 & 68.3 & 77.9 & 76.7 & 79.7 & 68.0 & 65.7 & 77.0 \\
& RocketEval & 67.9 & 76.9 & 74.8 & 75.6 & 57.9 & 63.4 & 70.8 & 73.5 & 71.6 & 66.5 & 64.9 & 74.2 \\
\rowcolor{lightblue}
& UCS (EN)   & 74.4 & 84.2 & 81.2 & 85.0 & 61.2 & 68.1 & 79.0 & 82.7 & 82.4 & 73.7 & 67.8 & 82.0 \\
%\rowcolor{lightblue}
%& Adaptive UCS (EN) & 72.8 & 83.0 & 79.6 & 83.1 & 58.9 & 66.3 & 82.2 & 79.8 & 80.4 & 71.0 & 68.9 & 79.9 \\
\midrule

% ================= Qwen 3 32B =================
\multirow{10}{*}{Qwen 3 32B}
& Zero-shot  & 74.9 & 73.7 & 71.4 & 78.5 & 56.6 & 63.5 & 71.9 & 72.3 & 69.2 & 65.0 & 67.9 & 76.0 \\
& CoT        & 71.7 & 73.8 & 72.6 & 77.2 & 55.2 & 62.9 & 72.0 & 72.4 & 70.0 & 66.7 & 68.1 & 76.5 \\
& AG         & 73.7 & 73.9 & 74.8 & 81.8 & 58.3 & 63.9 & 76.6 & 73.0 & 70.9 & 68.4 & 66.1 & 80.0 \\
& ChatEval   & 66.7 & 68.5 & 63.4 & 66.9 & 50.9 & 56.3 & 61.2 & 65.1 & 61.7 & 59.4 & 56.9 & 66.0 \\
& CheckEval  & 71.1 & 74.6 & 72.3 & 80.9 & 59.2 & 63.5 & 74.5 & 73.6 & 71.1 & 65.8 & 65.0 & 79.0 \\
& RocketEval & 70.0 & 72.9 & 68.7 & 72.4 & 54.8 & 59.7 & 65.8 & 69.0 & 66.9 & 63.9 & 61.4 & 70.4 \\
\rowcolor{lightblue}
& UCS (EN)   & 74.4 & 82.7 & 79.5 & 85.0 & 57.7 & 61.7 & 68.4 & 73.4 & 72.1 & 68.9 & 66.7 & 78.9 \\
%\rowcolor{lightblue}
%& Adaptive UCS (EN) & 73.6 & 80.9 & 76.8 & 83.1 & 56.1 & 63.9 & 67.0 & 70.4 & 69.6 & 66.4 & 65.9 & 76.9 \\
\midrule

% ================= Llama70b =================
\multirow{10}{*}{Llama70b}
& Zero-shot  & 69.9 & 70.3 & 68.0 & 73.8 & 55.6 & 57.9 & 65.3 & 73.2 & 60.4 & 63.4 & 62.8 & 75.8 \\
& CoT        & 67.1 & 72.2 & 68.4 & 69.7 & 56.6 & 58.2 & 64.8 & 75.3 & 60.7 & 64.7 & 62.3 & 75.8 \\
& AG         & 75.4 & 76.4 & 74.0 & 80.8 & 58.2 & 59.0 & 68.7 & 77.6 & 63.2 & 66.1 & 67.3 & 77.5 \\
& ChatEval   & 68.4 & 71.3 & 66.1 & 69.8 & 53.9 & 57.4 & 63.9 & 68.9 & 64.8 & 62.1 & 59.8 & 69.6 \\
& CheckEval  & 72.8 & 76.6 & 72.1 & 80.2 & 59.2 & 59.2 & 69.5 & 77.1 & 65.4 & 63.8 & 67.8 & 73.9 \\
& RocketEval & 71.8 & 74.9 & 70.5 & 73.9 & 57.1 & 61.0 & 67.9 & 72.3 & 69.5 & 66.3 & 64.0 & 73.1 \\
\rowcolor{lightblue}
& UCS (EN)   & 77.3 & 80.8 & 82.1 & 84.5 & 63.5 & 56.5 & 72.1 & 74.0 & 73.2 & 67.4 & 66.1 & 77.7 \\
%\rowcolor{lightblue}
%& Adaptive UCS (EN) & 73.6 & 80.9 & 76.8 & 83.1 & 56.1 & 63.9 & 67.0 & 70.4 & 69.6 & 66.4 & 65.9 & 76.9 \\
\midrule

% ================= OSS-120B =================
\multirow{8}{*}{OSS-120B}
& Zero-shot  
& 79.2 & 83.6 & 79.3 & 81.5
& 64.1 & 67.6 & 82.0 & 78.6 & 74.9 & 74.2 & 69.4 & 85.1 \\

& CoT        
& 78.3 & 82.9 & 78.8 & 80.9
& 64.7 & 67.0 & 81.4 & 79.4 & 74.1 & 73.9 & 69.9 & 83.8 \\

& AG         
& 79.4 & 84.1 & 79.0 & 82.1
& 65.2 & 66.7 & 82.6 & 79.7 & 75.0 & 74.6 & 69.7 & 85.3 \\

& ChatEval   
& 73.1 & 74.2 & 64.9 & 66.1
& 58.3 & 65.4 & 69.4 & 74.0 & 65.7 & 66.8 & 58.0 & 71.4 \\

& CheckEval  
& 78.6 & 83.8 & 78.5 & 81.9
& 65.7 & 67.9 & 82.1 & 79.1 & 75.2 & 75.0 & 69.1 & 84.7 \\

& RocketEval 
& 78.9 & 84.1 & 78.9 & 82.3
& 66.0 & 68.2 & 82.4 & 79.3 & 75.4 & 75.2 & 69.4 & 85.0 \\
\rowcolor{lightblue}
& UCS (EN)   
& 79.4 & 83.9 & 79.7 & 81.9
& 64.4 & 67.8 & 82.4 & 79.0 & 75.1 & 74.6 & 69.7 & 85.6 \\

%\rowcolor{lightblue}
%& Adaptive UCS (EN)   
%& 79.3 & 83.8 & 79.5 & 81.6
%& 64.1 & 67.5 & 82.4 & 78.7 & 74.9 & 75.0 & 69.6 & 85.9 \\

\bottomrule
\end{tabular}
\caption{Balanced Accuracy (BA) of all baselines and criteria set judges across languages on MEMERAG and mFACE. }
\label{app:fullcrosslingual}

\end{table*}
~\autoref{fig:correlation} shows the cross-lingual alignment of criterion importance between English and target languages for MEMERAG and mFACE. The heatmaps illustrate how the relative importance of evaluation criteria varies across languages.

\begin{figure*}[h]
\centering
\begin{tcolorbox}[
  title=Summary Evaluation Questions,
  width=\textwidth,
  colback=gray!3,
  colframe=black,
  boxrule=0.6pt,
  arc=2mm,
  left=1.5mm,right=1.5mm,top=1mm,bottom=1mm
]
\tiny
\textbf{Factual Accuracy (6 questions)}
\begin{enumerate}[label=\arabic*., leftmargin=*, itemsep=0.35em]
  \item Does the summary include any claims that are not explicitly supported by information in the source document?
  \item Does the summary contradict any specific fact, detail, or relationship stated in the source document?
  \item Are events, actions, or attributes in the summary correctly attributed to the individuals, entities, or sources named in the source document?
  \item Does the summary present speculative, uncertain, or conditional information from the source as definitive or certain?
  \item Does the summary accurately reflect the relative importance or prominence of key points as presented in the source document?
  \item Does the summary introduce numerical data, statistics, or quantitative claims that differ from those in the source document?
\end{enumerate}

\textbf{Contradiction Detection (6 questions)}
\begin{enumerate}[label=\arabic*., leftmargin=*, itemsep=0.35em, resume]
  \item Does the summary attribute a claim, opinion, or action to a source or entity that is not supported or explicitly stated in the source document?
  \item Does the summary present a possibility or uncertainty as a definitive fact, thereby increasing the level of certainty beyond what is expressed in the source document?
  \item Does the summary state that an event occurred, or a condition exists, when the source document explicitly indicates it did not happen or was not the case?
  \item Does the summary include a causal relationship between two events that is not stated, implied, or supported by the source document?
  \item Does the summary report a numerical value, statistic, or quantitative detail that contradicts the figures provided in the source document?
  \item Does the summary describe an action or outcome as having happened in the past when the source document states it is planned, proposed, or speculative?
\end{enumerate}

\textbf{Claim Support (6 questions)}
\begin{enumerate}[label=\arabic*., leftmargin=*, itemsep=0.35em, resume]
  \item Does the summary include any claims that are not explicitly supported by evidence or statements in the source document?
  \item Are there any statements in the summary that directly contradict information provided in the source document?
  \item Does the summary attribute actions, opinions, or statements to individuals or entities that are not assigned to them in the source document?
  \item To what extent does the summary present speculative or conditional information from the source as definitive or certain?
  \item Does the summary introduce new causal relationships or implications between events that are not stated or logically supported in the source document?
  \item Are key details in the summary distorted in a way that alters the meaning or significance of the information presented in the source document?
\end{enumerate}

\textbf{Misrepresentation Identification (6 questions)}
\begin{enumerate}[label=\arabic*., leftmargin=*, itemsep=0.35em, resume]
  \item Does the summary attribute a claim, opinion, or action to a source or entity that is not supported or explicitly stated in the source document?
  \item Does the summary present a speculative or conditional statement from the source as a definitive fact?
  \item Does the summary include a key event, outcome, or statistic that is not mentioned or implied in the source document?
  \item Does the summary reverse, invert, or otherwise distort the causal or temporal relationship between two events described in the source document?
  \item Does the summary exaggerate the strength, scope, or certainty of a finding, trend, or conclusion beyond what is stated in the source document?
  \item Does the summary omit a critical qualifying condition, limitation, or exception present in the source that changes the interpretation of the information?
\end{enumerate}

\textbf{Certainty Calibration (6 questions)}
\begin{enumerate}[label=\arabic*., leftmargin=*, itemsep=0.35em, resume]
  \item Does the summary present a claim as certain or definitive when the source document expresses it as uncertain, tentative, or conditional?
  \item Does the summary introduce a level of confidence or precision (e.g., ``proves,'' ``definitely,'' ``exactly'') in a claim that is not supported by the degree of certainty used in the source document?
  \item Are probabilities, frequencies, or likelihoods in the summary accurately reflected in terms of their magnitude and wording compared to the source document (e.g., ``likely'' vs.\ ``possible'' vs.\ ``certain'')?
  \item Does the summary omit hedging language (e.g., ``may,'' ``suggests,'' ``appears to'') present in the source, thereby overstating the strength of a conclusion?
  \item To what extent does the summary mirror the source document's attribution of claims to specific agents, studies, or evidence, without shifting responsibility or generalizing to broader consensus?
  \item Does the summary elevate a hypothesis, preliminary finding, or speculative idea from the source to the status of an established fact?
\end{enumerate}

\end{tcolorbox}
\label{app:questionmeme}
\end{figure*}

\begin{figure*}[h]
\centering
\begin{tcolorbox}[
  title=Evidence Support Evaluation Questions,
  width=\textwidth,
  colback=gray!3,
  colframe=black,
  boxrule=0.6pt,
  arc=2mm,
  fontupper=\footnotesize,
  left=1.5mm,right=1.5mm,top=1mm,bottom=1mm
]
\tiny
\textbf{Factual Accuracy (6 questions)}
\begin{enumerate}[label=\arabic*., leftmargin=*, itemsep=0.25em]
  \item To what extent does the evidence directly support the key claims in the answer, with no unsupported assertions?
  \item How well do the retrieved passages contradict or fail to support any statements in the answer?
  \item To what degree does the answer avoid overgeneralizing or making broad inferences beyond what is stated in the evidence?
  \item How precisely does the answer align with the specificity (e.g., timeframes, quantities, conditions) provided in the evidence?
  \item To what extent are all named entities, events, and relationships in the answer accurately reflected in the evidence passages?
  \item How well does the answer refrain from introducing plausible but unverified details not present in the evidence?
\end{enumerate}

\textbf{Completeness of Support (6 questions)}
\begin{enumerate}[label=\arabic*., leftmargin=*, itemsep=0.25em, resume]
  \item To what extent does the evidence directly support all key claims in the answer, with no unsupported assertions?
  \item How well do the retrieved passages contain specific details or examples that match the level of specificity in the answer?
  \item To what degree does the evidence fully cover the scope of the answer, including all sub-claims or components mentioned?
  \item How consistently do the passages align with the answer without introducing contradictions or conflicting information?
  \item To what extent does the answer avoid overgeneralizing beyond what is reasonably supported by the evidence?
  \item How well do the passages provide sufficient context or explanation to justify causal or inferential claims made in the answer?
\end{enumerate}

\textbf{Specificity Alignment (6 questions)}
\begin{enumerate}[label=\arabic*., leftmargin=*, itemsep=0.25em, resume]
  \item To what extent does the answer reflect the same level of specificity as the evidence, avoiding unwarranted generalizations or oversimplifications?
  \item How well do the key claims in the answer map directly to specific details or data points in the evidence, rather than relying on vague or peripheral information?
  \item To what degree does the evidence support the precise scope (e.g., time frame, population, location) asserted in the answer without overextension?
  \item How closely does the answer avoid introducing concepts or conclusions that are more specific than what is warranted by the evidence?
  \item To what extent are named entities, quantities, or relationships in the answer explicitly grounded in corresponding specific mentions within the evidence?
  \item How well does the answer maintain alignment with the evidence by neither omitting critical qualifying conditions nor adding unsupported qualifiers?
\end{enumerate}

\textbf{Consistency with Evidence (6 questions)}
\begin{enumerate}[label=\arabic*., leftmargin=*, itemsep=0.25em, resume]
  \item To what extent does the evidence directly support the key claims in the answer, with no unsupported assertions?
  \item How well do the retrieved passages contradict or conflict with any statements in the answer?
  \item To what degree does the answer avoid overgeneralizing beyond the scope or specificity of the evidence provided?
  \item How closely does the answer align with the factual details and context present in the evidence, avoiding subtle distortions or misrepresentations?
  \item To what extent can each component of a multi-part answer be individually justified by at least one evidence passage?
  \item How well does the answer reflect the certainty level (e.g., tentative, definitive) expressed in the evidence, without introducing unwarranted confidence or ambiguity?
\end{enumerate}

\textbf{Source Attribution (6 questions)}
\begin{enumerate}[label=\arabic*., leftmargin=*, itemsep=0.25em, resume]
  \item To what extent does the evidence explicitly attribute the claim to a credible source or original provider of information?
  \item How well do the retrieved passages support the specificity of the claim, without introducing unsupported details or omitting critical qualifiers present in the source?
  \item To what degree is the claim directly verifiable from the cited evidence, rather than requiring inference beyond what the source states?
  \item How consistently does the answer reflect the source's intended meaning, avoiding misrepresentation or overgeneralization of the evidence?
  \item To what extent does the evidence rule out contradictions or significant discrepancies with the claim being made?
  \item How clearly is the connection between the evidence and the claim articulated, such that the support is traceable and transparent?
\end{enumerate}

\end{tcolorbox}
\label{app:questionmFACE}

\end{figure*}

~\autoref{tab:T1_3run_diverse_memerag} and ~\autoref{tab:T1_3run_diverse_mFACE} report MEMERAG and mFACE results for Qwen-235B averaged over three runs with different seeds. The results show that UCS maintains consistent improvements across languages while exhibiting relatively low variance compared to most baselines.

\begin{table*}[t]
\centering
\tiny
\begin{tabular}{lcccc}
\toprule
Method & DE & FR & ES & HI \\
\midrule
Zero-shot  & 77.2$\pm$0.4 & 77.5$\pm$1.3 & 77.5$\pm$0.9 & 74.4$\pm$1.1 \\
CoT        & 73.0$\pm$1.2 & 76.2$\pm$0.5 & 76.2$\pm$1.4 & 74.1$\pm$0.8 \\
AG         & 77.2$\pm$0.7 & 80.6$\pm$1.1 & 76.1$\pm$0.4 & 78.0$\pm$1.4 \\
ChatEval   & 71.8$\pm$1.5 & 73.0$\pm$0.6 & 60.9$\pm$1.9 & 62.4$\pm$0.7 \\
CheckEval  & 76.1$\pm$0.5 & 80.7$\pm$1.4 & 74.8$\pm$0.7 & 78.4$\pm$1.1 \\
RocketEval & 76.4$\pm$1.9 & 81.0$\pm$1.2 & 75.9$\pm$1.7 & 79.1$\pm$1.4 \\
UCS (EN)   & \textbf{79.7}$\pm$0.6 & \textbf{84.3}$\pm$1.1 & \textbf{80.0}$\pm$0.4 & \textbf{82.2}$\pm$1.2 \\
\bottomrule
\end{tabular}
\caption{Balanced Accuracy (BA) on MEMERAG for Qwen-235B, reported as mean $\pm$ standard deviation over 3 runs.}
\label{tab:T1_3run_diverse_memerag}
\end{table*}

\begin{table*}[t]
\centering
\tiny
\begin{tabular}{lcccccccc}
\toprule
Method & AM & MY & FR & SW & TH & AR & HI & ES \\
\midrule

Zero-shot  & 54.7$\pm$0.7 & \textbf{70.2}$\pm$1.3 & 74.3$\pm$0.6 & 73.0$\pm$1.1 & 72.6$\pm$0.4 & 69.7$\pm$1.4 & 68.0$\pm$1.0 & 77.9$\pm$0.6 \\
CoT        & 58.7$\pm$1.4 & 68.5$\pm$0.6 & 72.8$\pm$1.2 & 77.5$\pm$0.5 & 71.4$\pm$1.1 & 69.2$\pm$0.5 & 69.4$\pm$1.6 & 74.4$\pm$0.8 \\
AG         & 59.3$\pm$0.8 & 67.4$\pm$1.5 & 76.6$\pm$0.7 & 77.7$\pm$1.1 & 72.6$\pm$0.6 & 70.7$\pm$1.2 & 69.0$\pm$0.7 & 79.5$\pm$1.2 \\
ChatEval   & 51.1$\pm$1.6 & 65.5$\pm$0.9 & 68.4$\pm$1.3 & 74.0$\pm$0.5 & 63.9$\pm$1.8 & 66.1$\pm$0.6 & 55.7$\pm$1.1 & 70.3$\pm$1.5 \\
CheckEval  & 60.8$\pm$1.3 & 68.9$\pm$0.4 & 74.9$\pm$1.6 & 76.8$\pm$0.7 & \textbf{75.1}$\pm$1.1 & 73.4$\pm$0.8 & 65.8$\pm$1.5 & 76.4$\pm$0.6 \\
RocketEval & 61.2$\pm$2.1 & 69.2$\pm$1.1 & 75.4$\pm$1.7 & 77.0$\pm$1.3 & 74.9$\pm$1.6 & 72.9$\pm$2.0 & 66.9$\pm$1.4 & 78.1$\pm$1.8 \\
UCS (EN)   & \textbf{64.7}$\pm$1.0 & 67.7$\pm$1.4 & \textbf{82.4}$\pm$0.6 & \textbf{79.0}$\pm$1.3 & \textbf{75.1}$\pm$0.4 & \textbf{74.6}$\pm$1.2 & \textbf{70.0}$\pm$0.7 & \textbf{85.5}$\pm$1.0 \\
\bottomrule
\end{tabular}
\caption{Balanced Accuracy (BA) on mFACE for Qwen-235B, reported as mean $\pm$ standard deviation over 3 runs.}
\label{tab:T1_3run_diverse_mFACE}
\end{table*}

~\autoref{app:thresholdagg} compares the transfer-module classifier with LLM-based aggregation of the criteria representations across languages. The trained transfer module generally achieves higher performance, although LLM aggregation remains competitive in some cases.

\begin{table*}[h]
\small
\centering
\begin{tabular}{llcc}
\toprule
Dataset & Lang. & UCS (transfer) & UCS (LLM) \\
\midrule
\multirow{4}{*}{MEMERAG}
 & DE & \textbf{79.7} & 76.9 \\
 & FR & \textbf{84.3} & 78.9 \\
 & ES & \textbf{80.0} & 77.2 \\
 & HI & \textbf{82.2} & 75.9 \\
\midrule
\multirow{8}{*}{mFACE}
 & AM & \textbf{64.7} & 62.5 \\
 & MY & 67.7 & \textbf{69.6} \\
 & FR & \textbf{82.4} & 82.3 \\
 & SW & \textbf{79.0} & 78.4 \\
 & TH & \textbf{75.1} & 74.9 \\
 & AR & \textbf{74.6} & 73.1 \\
 & HI & \textbf{70.0} & 68.8 \\
 & ES & \textbf{85.5} & 84.5 \\
\bottomrule
\end{tabular}
\caption{Comparison of threshold-based training classifier and LLM-based aggregation of intermediate criteria representations for final judgment.}
\label{app:thresholdagg}
\end{table*}

\section{Prompts}
\label{app:prompts}

\begin{figure*}[h]
\centering
\begin{tcolorbox}[
  title=Concept Generation - MEMERAG,
  width=\textwidth,
  colback=gray!3,
  colframe=black,
  boxrule=0.6pt,
  arc=2mm,
  fontupper=\footnotesize,
  left=1.5mm,right=1.5mm,top=1mm,bottom=1mm
]

\textbf{}

\vspace{2mm}

\noindent\texttt{"system\_prompt": "You are an evidence-grounding evaluator. Your goal is to identify key verification concepts for checking whether answers are supported by evidence."}

\vspace{2mm}

\noindent\texttt{"task\_prompt": "Generate 3–5 distinct verification concepts that are essential for evaluating whether an answer is supported by evidence passages.\textbackslash n\textbackslash nGuidelines:\textbackslash n- Concepts should cover different aspects of evidence grounding (e.g., factual accuracy, completeness, specificity, consistency, source attribution).\textbackslash n- Each concept should be general enough to apply across different topics and domains.\textbackslash n- Concepts should target different failure modes where answers might not be properly supported.\textbackslash n- Keep concepts concise (2–5 words each).\textbackslash n\textbackslash nReturn exactly this format:\textbackslash n\textbackslash n<concepts>\textbackslash n<concept1>[Concept name]</concept1>\textbackslash n<concept2>[Concept name]</concept2>\textbackslash n...\textbackslash n</concepts>"}

\end{tcolorbox}
\end{figure*}

\begin{figure*}[h]
\centering
\begin{tcolorbox}[
  title=Question Generation - MEMERAG,
  width=\textwidth,
  colback=gray!3,
  colframe=black,
  boxrule=0.6pt,
  arc=2mm,
  fontupper=\footnotesize,
  left=1.5mm,right=1.5mm,top=1mm,bottom=1mm
]

\textbf{}

\vspace{2mm}

\noindent\texttt{"system\_prompt": "You are an evidence-grounding evaluator. Your goal is to generate reusable verification questions for checking whether an answer is supported by retrieved evidence passages."}

\vspace{2mm}

\noindent\texttt{"task\_prompt": "**Verification Concept:** {concept}Generate 4–6 clear, diverse, and reusable evaluation questions for verifying whether an answer is supported by evidence with respect to the concept '{concept}'. These questions will later be scored on a **0–10 scale indicating degree of evidence support**.Guidelines:- Focus on whether an **answer is supported by evidence passages**.- Questions must be reusable across different topics and documents.- Cover different **evidence-grounding failure modes** (e.g., unsupported claims, contradictions, overgeneralization, specificity mismatch).- Each question should be written to support **graded (partial-to-full) scoring**, not just yes/no judgments.- Each question should target a distinct verification angle.- Wording should be concise and evaluative (e.g., \"To what extent is the claim supported by the evidence…\", \"How well do the passages justify…\").Return exactly this format:<questions><question1>Question: [Universal evaluation question]</question1>...</questions>"}

\end{tcolorbox}
\end{figure*}

\begin{figure*}[h]
\centering
\begin{tcolorbox}[
  title=Question Application - MEMERAG,
  width=\textwidth,
  colback=gray!3,
  colframe=black,
  boxrule=0.6pt,
  arc=2mm,
  fontupper=\footnotesize,
  left=1.5mm,right=1.5mm,top=1mm,bottom=1mm
]

\textbf{}

\vspace{2mm}

\noindent\texttt{"system\_prompt": "You are an evidence-grounding evaluator. Your task is to determine whether an answer is supported by the provided evidence passages."}

\vspace{2mm}

\noindent\texttt{
"task\_prompt": "**Evidence Passages:**\textbackslash n
\{context\}\textbackslash n\textbackslash n
**Answer:** \{answer\_segment\}\textbackslash n
**Verification Concept:** \{concept\}\textbackslash n\textbackslash n
**Evaluation Questions:**\textbackslash n
\{questions\}\textbackslash n\textbackslash n
For each question, assign a score from **0--10** indicating how well the evidence supports the answer with respect to that question.\textbackslash n\textbackslash n
Base all judgments strictly on the provided evidence passages. Do not assume any external knowledge.\textbackslash n\textbackslash n
Return exactly this format:\textbackslash n\textbackslash n
<evaluation>\textbackslash n
\{questions\_format\_example\}\textbackslash n
</evaluation>"
}
\end{tcolorbox}
\end{figure*}

\begin{figure*}[t]
\centering

% -------------------- Box 1 --------------------
\begin{tcolorbox}[
  title=Concept Generation - mFACE,
  width=\textwidth,
  colback=gray!3,
  colframe=black,
  boxrule=0.6pt,
  arc=2mm,
  fontupper=\footnotesize,
  left=1.5mm,right=1.5mm,top=1mm,bottom=1mm
]
\small
\textbf{concept\_generation} \\[1mm]
\texttt{"system\_prompt": "You are an expert factual faithfulness evaluator. Your goal is to identify key faithfulness concepts for checking whether summaries are factually faithful to their source documents."} \\[1mm]
\texttt{"task\_prompt": "Generate 3–5 distinct faithfulness concepts that are essential for evaluating whether a summary is factually faithful to its source document.\textbackslash n\textbackslash nGuidelines:\textbackslash n- Concepts should cover different aspects of factual faithfulness (e.g., factual accuracy, contradiction detection, unsupported claims, misrepresentation, distorted certainty).\textbackslash n- Each concept should be general enough to apply across different topics and domains.\textbackslash n- Concepts should target different failure modes where summaries might not be properly supported by source documents.\textbackslash n- Keep concepts concise (2–5 words each).\textbackslash n\textbackslash nReturn exactly this format:\textbackslash n\textbackslash n<concepts>\textbackslash n<concept1>[Concept name]</concept1>\textbackslash n<concept2>[Concept name]</concept2>\textbackslash n...\textbackslash n</concepts>"} 
\end{tcolorbox}

\vspace{2mm}

% -------------------- Box 2 --------------------
\begin{tcolorbox}[
  title=Question Generation - mFACE,
  width=\textwidth,
  colback=gray!3,
  colframe=black,
  boxrule=0.6pt,
  arc=2mm,
  fontupper=\footnotesize,
  left=1.5mm,right=1.5mm,top=1mm,bottom=1mm
]
\small
\textbf{question\_generation} \\[1mm]
\texttt{"system\_prompt": "You are an expert factual faithfulness evaluator. Your role is to construct high-quality, reusable evaluation questions that test whether summaries are factually faithful to their source documents."} \\[1mm]
\texttt{"task\_prompt": "**Faithfulness Concept:** \{concept\}\textbackslash n\textbackslash nGenerate exactly 6 clear, diverse, and reusable evaluation questions for the faithfulness concept '\{concept\}'. Each question must require direct comparison between a summary and its source document to assess factual support, contradiction, or distortion under this concept.\textbackslash n\textbackslash nAnnotation Guidance to Apply While Writing Questions:\textbackslash n- The question must be answerable ONLY by checking the source document.\textbackslash n- The question should detect one of the following: unsupported claims, contradiction, misrepresentation, distorted certainty, or incorrect attribution.\textbackslash n- The question must not depend on surface form (grammar, fluency, style, or verbosity).\textbackslash n\textbackslash nGuidelines:\textbackslash n- Every question must explicitly or implicitly require verification against the source document.\textbackslash n- Do NOT ask about readability, grammar, fluency, writing quality, or length.\textbackslash n- Questions must be context-independent and reusable across domains.\textbackslash n- All questions must stay strictly within the given faithfulness concept, but vary the factual scenario or common error pattern being tested.\textbackslash n- Avoid redundancy: each question should target a distinct factual failure mode.\textbackslash n- Wording must be concise, unambiguous, and evaluative (e.g., \textbackslash "Does the summary…\textbackslash " or \textbackslash "To what extent does the summary…\textbackslash ").\textbackslash n\textbackslash nReturn exactly this format:\textbackslash n\textbackslash n<questions>\textbackslash n<question1>\textbackslash nQuestion: [Universal evaluation question]\textbackslash n</question1>\textbackslash n\textbackslash n<question2>\textbackslash nQuestion: [Universal evaluation question]\textbackslash n</question2>\textbackslash n\textbackslash n<question3>\textbackslash nQuestion: [Universal evaluation question]\textbackslash n</question3>\textbackslash n\textbackslash n<question4>\textbackslash nQuestion: [Universal evaluation question]\textbackslash n</question4>\textbackslash n\textbackslash n<question5>\textbackslash nQuestion: [Universal evaluation question]\textbackslash n</question5>\textbackslash n\textbackslash n<question6>\textbackslash nQuestion: [Universal evaluation question]\textbackslash n</question6>\textbackslash n</questions>"} 
\end{tcolorbox}

\vspace{2mm}

% -------------------- Box 3 --------------------
\begin{tcolorbox}[
  title=Question Application - mFACE,
  width=\textwidth,
  colback=gray!3,
  colframe=black,
  boxrule=0.6pt,
  arc=2mm,
  fontupper=\footnotesize,
  left=1.5mm,right=1.5mm,top=1mm,bottom=1mm
]
\small
\textbf{question\_application} \\[1mm]
\texttt{"system\_prompt": "You are an expert factual faithfulness evaluator. Your task is to score a summary against its source document using predefined faithfulness evaluation questions."} \\[1mm]
\texttt{"task\_prompt": "**Source Document:**\textbackslash n\{context\}\textbackslash n\textbackslash n**Summary:** \{answer\_segment\}\textbackslash n\textbackslash n**Faithfulness Concept:** \{concept\}\textbackslash n\textbackslash n**Evaluation Questions:**\textbackslash n\{questions\}\textbackslash n\textbackslash nFor each question, evaluate whether the summary is factually faithful to the source document under that question.\textbackslash n\textbackslash nWhile scoring, explicitly consider whether the summary:\textbackslash n- Contains information not stated in or directly inferable from the source\textbackslash n- Contradicts the source\textbackslash n- Introduces unsupported details\textbackslash n- Misrepresents entities, quantities, relationships, or certainty\textbackslash n- Overgeneralizes, narrows, or distorts scope\textbackslash n\textbackslash nAssign a score from 0–10 for each question:\textbackslash n- 0 = Completely unfaithful (clear contradiction, fabrication, or unsupported claim)\textbackslash n- 10 = Completely faithful (fully supported, correctly represented, no distortion)\textbackslash n\textbackslash nRules:\textbackslash n- Base every score strictly on evidence from the source document.\textbackslash n- Ignore grammar, fluency, style, and summary length.\textbackslash n- Penalize both direct hallucinations and subtle distortions of meaning or certainty.\textbackslash n\textbackslash nProvide a brief, evidence-based justification for each score.\textbackslash n\textbackslash nReturn exactly this format:\textbackslash n\textbackslash n<evaluation>\textbackslash n\{questions\_format\_example\}\textbackslash n</evaluation>"} 
\end{tcolorbox}

\end{figure*}

\end{document}